\documentclass{article}
\usepackage{ijcai21}

\usepackage{times}
\usepackage{soul}
\usepackage{url}
\usepackage[hidelinks]{hyperref}  
\usepackage[utf8]{inputenc}
\usepackage[small]{caption}
\usepackage{graphicx}
\usepackage{amsmath}
\usepackage{amsthm}
\usepackage{booktabs}
\usepackage{algorithm}
\usepackage{algorithmic}
\usepackage{subcaption}
\usepackage{balance}

\title{``Why Would I Trust Your Numbers?''\\On the Explainability of Expected Values in Soccer}

\author{
    Jan Van Haaren
    \affiliations
    KU Leuven, Belgium
    \emails
    jan.vanhaaren@kuleuven.be
}

\begin{document}

\maketitle

\begin{abstract}
In recent years, many different approaches have been proposed to quantify the performances of soccer players. Since player performances are challenging to quantify directly due to the low-scoring nature of soccer, most approaches estimate the expected impact of the players' on-the-ball actions on the scoreline. While effective, these approaches are yet to be widely embraced by soccer practitioners. The soccer analytics community has primarily focused on improving the accuracy of the models, while the \textit{explainability} of the produced metrics is often much more important to practitioners.

To help bridge the gap between scientists and practitioners, we introduce an \textit{explainable} Generalized Additive Model that estimates the expected value for shots. Unlike existing models, our model leverages features corresponding to widespread soccer concepts. To this end, we represent the locations of shots by fuzzily assigning the shots to designated zones on the pitch that practitioners are familiar with. Our experimental evaluation shows that our model is as accurate as existing models, while being easier to explain to soccer practitioners. 
\end{abstract}

\section{Introduction}

Due to the low-scoring nature of soccer, the performances of individual soccer players are challenging to appropriately quantify. Most actions that players perform in a match do not impact the scoreline directly but they may affect the course of the match. Until the mid 2010s, soccer clubs primarily relied on count-based statistics for match analysis and player recruitment purposes. These statistics typically focus on events that are easy to count such as the number of completed passes, number of shots on target or number of tackles. However, they are typically not representative of the players' actual performances as they often fail to account for the relevant context. Most traditional statistics treat each action of a given type as equally valuable, which is never the case in practice.

The increased availability of detailed match event data has enabled more sophisticated approaches to quantify the performances of soccer players in matches. These approaches exploit the observation that the outcome of a particular action depends on many different factors, many of which are often beyond the power of the player who executes the action. For instance, a line-breaking pass in midfield is extremely valuable if the teammate at the receiving end of the pass manages to put the ball in the back of the net, or can be perceived as useless if the same teammate fumbles their first touch.

The majority of the performance metrics that are currently used in soccer rely on the expected value of the actions that the players perform in matches. Ignoring the actual impact of the actions, these approaches reward each action based on the impact that highly similar actions had in the past~\cite{decroos2017starss,decroos2019actions,vanroy2020valuing}. As a result, a particular type of action that led to a goal further down the line several times in the past will be deemed more valuable than a particular type of action that has never led to a goal before. The value of an action is typically expressed in terms of the action's impact on their team's likelihood of scoring and/or conceding a goal within a given time period or number of actions after the valued action. For example, the VAEP framework uses a ten-action look-ahead window to value an action in terms of likelihood of scoring or preventing a goal~\cite{decroos2019actions}.

Although many different approaches to valuing player actions have been proposed in recent years~\cite{rudd2011framework,decroos2017starss,vanroy2020valuing,liu2020deep}, the adoption of these approaches by soccer practitioners has been quite limited to date. Soccer practitioners such as performance analysts, recruitment analysts and scouts are often reluctant to base their decisions on performance metrics that provide little insights into how the numbers came about. The community's efforts have primarily gone to devising alternative methods and improving the accuracy of existing methods.

Although the \textit{explainability} of the expected value for a particular action has received little to no attention to date, there has been some work on improving the \textit{interpretability} of the underlying models that produce the expected values. \cite{decroos2019interpretable} introduce an alternative implementation of the VAEP framework. Their approach differs from the original approach in two ways. First, they replace the Gradient Boosted Decision Tree models with Explainable Boosting Machine models, which combine the concepts of Generalized Additive Models and boosting. Second, they reduce the number of features from over 150 to only 10. Their implementation yields similar performance to the original implementation, but the models are easier to interpret. However, the resulting expected values for specific actions often remain hard to explain to soccer practitioners. The features were designed to facilitate the learning process and not to be explained to soccer practitioners who are unfamiliar with mathematics.

In an attempt to improve the \textit{explainability} of the expected values that are produced by the \cite{decroos2019interpretable} approach, this paper proposes the following two changes:
\begin{enumerate}
    \item The spatial location of each action is represented as a feature vector that reflects a fuzzy assignment to 16 pitch zones, which were inspired by the 12 pitch zones introduced in~\cite{caley2013shot}. The zones are meaningful to soccer practitioners as they are often used to convey tactical concepts and insights about the game~\cite{maric2014halfspaces}. In contrast, \cite{decroos2019interpretable} represent each spatial location using the distance to goal and the angle to goal like the original VAEP implementation.
    \item The potential interactions between two features are manually defined such that the learning algorithm can only leverage feature interactions that are meaningful from a soccer perspective when learning the model. In contrast, \cite{decroos2019interpretable} allow the learning algorithm to learn at most three arbitrary pairwise interactions, which do not necessarily capture concepts that are meaningful to soccer practitioners.
\end{enumerate}

The remainder of this paper is organized as follows. Section~\ref{sec:problem-statement} formally introduces the task at hand and Section~\ref{sec:dataset} describes the dataset. Section~\ref{sec:approach} introduces our approach to learning an \textit{explainable} model to estimate the expected values for shots. Section~\ref{sec:experimental-evaluation} provides an empirical evaluation of our approach. Section~\ref{sec:related-work} discusses related work. Section~\ref{sec:conclusion-future-work} presents the conclusions and avenues for further research.

\section{Problem Statement}
\label{sec:problem-statement}

This paper addresses the task of estimating expected values for shots, which are often referred to as ``expected-goals values'' or ``xG values'' in the soccer analytics community. The goal is to learn a model whose estimates can conveniently be explained to practitioners using the jargon that they are familiar with. More formally, the task at hand is defined as follows:

\begin{description}
    \item[Given:] characteristics of the shot (e.g., spatial location, body part used), and characteristics and outcomes of shots in earlier matches;
    \item[Estimate:] the probability of the shot yielding a goal;
    \item[Such that:] the estimate can be explained to soccer practitioners who are unfamiliar with machine learning.
\end{description}

Unlike this paper, \cite{decroos2019interpretable} addresses the broader task of estimating the likelihood of scoring and conceding a goal from a particular game state within the next ten actions. However, from a machine learning perspective, both tasks are very similar. In our case, the label depends on the outcome of the shot. In their case, the label depends on the outcome of potential shots within the following ten actions.

\section{Dataset}
\label{sec:dataset}

We use match event data that was collected from the Goal.com website for the top-5 European competitions: the English Premier League, German Bundesliga, Spanish LaLiga, Italian Serie A and French Ligue 1. Our dataset covers all matches in the 2017/2018 through 2019/2020 seasons as well as the 2020/2021 season until Sunday 2 May 2021. 

Our dataset describes each on-the-ball action that happened in each covered match. For each action, our dataset includes, among others, the type of the action (e.g., pass, shot, tackle, interception), the spatial location of the action (i.e., the x and y coordinate), the time elapsed since the start of the match, the body part that the player used to perform the action, the identity of the player who performed the action and the team that the player who performed the action plays for.

Table~\ref{tbl:dataset} shows the number of shots in each season in each competition in our dataset.

\begin{table}[htb]
    \centering
    \begin{tabular}{lrrrr}
    \toprule
    \textbf{Competition} & \textbf{2017/18} & \textbf{2018/19} & \textbf{2019/20} & \textbf{2020/21} \\
    \midrule
    Premier League & 9,328 & 9,667 & 9,429 & 5,893 \\
    Bundesliga & 7,760 & 8,273 & 8,152 & 4,849 \\
    LaLiga & 9,191 & 9,273 & 8,615 & 5,054 \\
    Serie A & 9,835 & 10,601 & 10,910 & 5,679 \\
    Ligue 1 & 9,475 & 9,428 & 6,829 & 6,182 \\
    \midrule
    \textbf{Total} & 45,592 & 47,242 & 43,935 & 38,737 \\
    \bottomrule
    \end{tabular}
    \caption{Number of shots in each season in each competition in our dataset. The 2020/2021 season is covered until Sunday 2 May 2021 when the last three to five rounds of the season still had to be played, depending on the competition.}
    \label{tbl:dataset}
\end{table}

\section{Approach}
\label{sec:approach}

We train a probabilistic classifier that estimates the probability of a shot resulting in a goal based on the characteristics of the shot. We first define a feature vector that reflects the shot characteristics and then pick a suitable model class.

\subsection{Representing Shot Characteristics}
\label{subsec:approach-shot-characteristics}

The location of the shot and the body part used by the player who performs the shot are known to be predictive of the outcome of a shot~\cite{caley2013shot,ijtsma2013forget,ijtsma2013where,caley2015premier}. Since the precise spatial locations of the players are unavailable, most approaches attempt to define features that are a proxy for the relevant context. For example, some approaches use the speed of play (e.g., distance covered by the ball within the last X seconds) as a proxy for how well the opposition's defense was organized at the time of the shot. That is, a defense tends to be less-well organized after a fast counter-attack than after a slow build-up.

We leverage a subset of the features that are used by the VAEP framework~\cite{decroos2019actions}. We process the data using the \textsc{socceraction} Python package.\footnote{\url{https://github.com/ML-KULeuven/socceraction}} First, we convert our dataset into the SPADL representation, which is a unified representation for on-the-ball player actions in soccer matches that facilitates analysis. Second, we produce the features that are used by the VAEP framework to represent game states. For our task, we only consider the game states from which a shot is taken. However, many VAEP features contribute little to nothing to the predictive power of the model while increasing the complexity of the model and the likeliness of over-fitting, as shown by~\cite{decroos2019interpretable}. Therefore, we only use the following four VAEP features:
\begin{itemize}
    \item \textbf{bodypart\_foot}: an indicator that is \textit{true} for footed shots and \textit{false} for all other shots;
    \item \textbf{bodypart\_head}: an indicator that is \textit{true} for headed shots and \textit{false} for all other shots;
    \item \textbf{bodypart\_other}: an indicator that is \textit{true} for non-footed and non-headed shots and \textit{false} for all other shots;
    \item \textbf{type\_shot\_penalty}: an indicator that is \textit{true} for penalty shots and \textit{false} for non-penalty shots.
\end{itemize}

Unlike~\cite{decroos2019interpretable}, we do not use VAEP's \textbf{start\_dist\_to\_goal} and \textbf{start\_angle\_to\_goal} features that represent the location of a shot as the distance and angle to the center of the goal, respectively. Instead, we use a zone-based representation for pitch locations.

\subsection{Representing Pitch Locations}
\label{subsec:approach-pitch-locations}

The location of a shot is an important predictor of the outcome of the shot. Hence, several different approaches to representing shot locations have been explored in the soccer analytics community~\cite{robberechts2020interplay}. However, the focus has been on optimizing the accuracy of the resulting models rather than on improving the \textit{explainability} of the representation to soccer practitioners. Shot locations are often represented as a combination of the distance and angle of the shot to the center of the goal. Although this representation allows machine learning algorithms to learn more accurate models, the representation is challenging to explain to soccer practitioners who are unfamiliar with mathematics.

We propose an alternative representation for shot locations that allows to learn accurate predictive models that can be more easily explained to soccer practitioners. We exploit the observation that soccer practitioners often refer to designated zones on the pitch to convey tactical concepts and insights about the game~\cite{maric2014halfspaces}. Hence, shot locations could be represented by overlaying the pitch with a grid and assigning each shot to the corresponding grid cell. However, shots that were taken from roughly the same location near a cell boundary may end up in different grid cells. As a result, these shots may look more dissimilar than they are from the perspective of the machine learning algorithm due to the \textit{hard} zone boundaries~\cite{decroos2020soccermix}.

\begin{figure}[h]
    \centering
    \includegraphics[width=\linewidth]{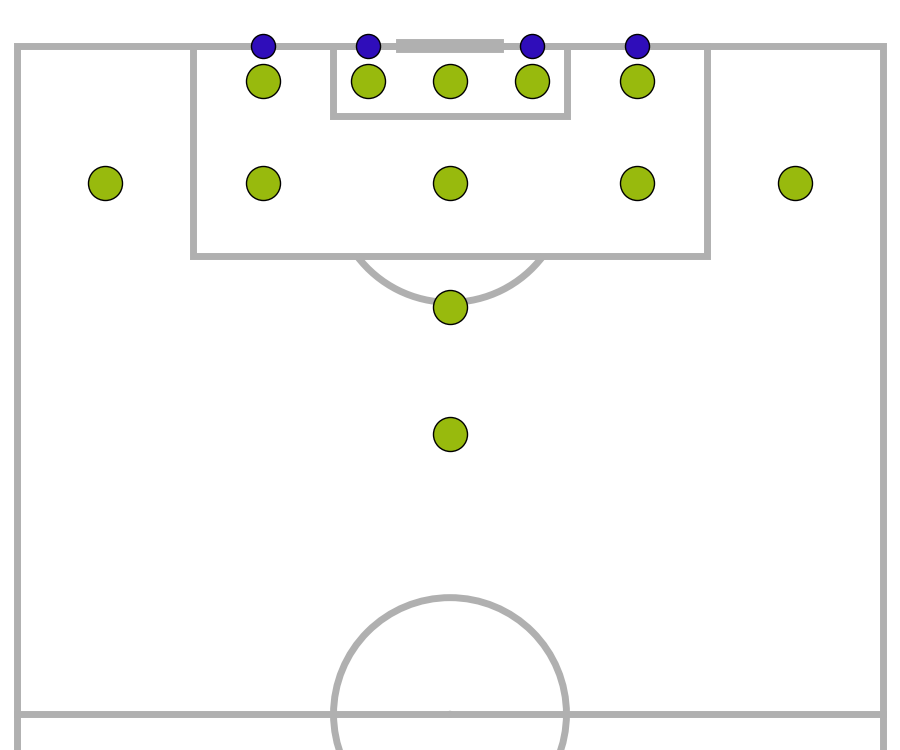}
    \caption{Definition of the zones. The circles show the centers of the 16 defined zones. The 12 green dots represent the zones identified by~\protect\cite{caley2013shot}. The additional 4 blue dots cover the goal line to improve the estimates for shots from a very tight angle.}
    \label{fig:zones}
\end{figure}

To overcome the limitations of \textit{hard} zone boundaries, we introduce an alternative representation that uses \textit{soft} zone boundaries. We perform a fuzzy assignment of shot locations to grid cells. That is, each shot location is assigned to a grid cell with a certain probability, where the probabilities sum to one across the grid. We perform c-means clustering~\cite{dunn1973fuzzy,bezdek1981pattern} using the \textsc{scikit-fuzzy} Python package.\footnote{\url{https://github.com/scikit-fuzzy/scikit-fuzzy}} First, we manually initialize the cluster centers, where each cluster center corresponds to the center of one of 16 zones, as shown in Figure~\ref{fig:zones}. The 12 green dots represent the zones identified by~\protect\cite{caley2013shot}. The additional 4 blue dots cover the goal line to improve the estimates for shots from very tight angles. Second, we run 1000 iterations of the c-means clustering algorithm to obtain the membership of each shot location to each of the clusters. We use the default value of 1000 iterations since running fewer or more iterations does not seem to affect the results. We also use the default exponent of 2 for the membership function. Smaller exponents lead to a \textit{too harsh} assignment of shot locations to clusters, whereas larger exponents lead to a \textit{too lenient} assignment such that valuable contextual information is lost.

\subsection{Training a Probabilistic Classifier}
\label{subsec:approach-classifier}

To estimate the probability of a shot yielding a goal, we train a probabilistic classifier. Like \cite{decroos2019interpretable}, we train an Explainable Boosting Machine. As the name suggests, the predictions made by Explainable Boosting Machine models can more easily be explained to soccer practitioners than the predictions made by traditional Gradient Boosted Decision Tree models, which are increasingly used in soccer analytics (e.g.,~\cite{decroos2019actions,bransen2019choke,bransen2020player}).

An Explainable Boosting Machine \cite{nori2019interpretml} is a Generalized Additive Model of the following form:

\begin{align*}
g(E[y]) = \beta_0 + \sum f_j(x_j) + \sum f_{ij}(x_i,x_j),
\end{align*}

where $g$ is the \textit{logit} link function in the case of classification, $y$ is the target, $\beta_0$ is the intercept, $f_j$ is a feature function that captures a feature $x_j$ and $f_{ij}$ is a feature function that captures the interaction between features $x_i$ and $x_j$. Each feature function is learned in turn by applying boosting and bagging on one feature or pair of features at a time. Due to the additive nature of the model, the contribution of each individual feature or feature interaction can be explained by inspecting or visualizing the feature functions $f_j$ and $f_{ij}$.

We train an Explainable Boosting Machine model using the \textsc{InterpretML} Python package.\footnote{\url{https://github.com/interpretml/interpret}} Unlike \cite{decroos2019interpretable}, who allow the algorithm to learn up to three arbitrary pairwise feature functions, we manually specify the pairwise feature functions that the algorithm is allowed to learn in order to guarantee the explainability of the feature interactions. We allow the algorithm to learn 25 pairwise feature functions: 12 feature functions that capture the relationship between each zone inside the penalty area and the footed-shot indicator, 12 feature functions that capture the relationship between each zone inside the penalty area and the headed-shot indicator, and a feature function that captures the relationship between the zone that contains the penalty spot and the penalty-shot indicator. Since headed shots from outside the penalty area are rare, the zones outside the penalty area are not considered in the pairwise feature functions.

\section{Experimental Evaluation}
\label{sec:experimental-evaluation}

We first discuss our experimental setup, then compare our approach to three baseline approaches, and finally analyze the explanations of the estimates made by our approach.

\subsection{Methodology}
\label{subsec:experimental-evaluation-methodology}

We compare our proposed zone-based approach to estimating the probability of a shot yielding a goal to three baseline approaches. Concretely, we consider the following four approaches in our experimental evaluation:
\begin{description}
    \item[Soft zones:] Our approach as introduced in Section~\ref{sec:approach};
    \item[Hard zones:] An alternative version of our approach where the \textit{soft} assignment of shot locations to zones is replaced by a \textit{hard} assignment. To keep the implementation as close as possible to the implementation of our original approach, we use an exponent of 1.001 for the membership function in the c-means clustering step;
    \item[Distance and angle:] An alternative version of our approach, inspired by \cite{decroos2019interpretable}, where the 16 features reflecting the \textit{soft} assignment of shot locations to zones is replaced by VAEP's \textbf{start\_dist\_to\_goal} and \textbf{start\_angle\_to\_goal} features that capture the shot's distance and angle to the center of the goal;
    \item[Naive baseline:] A simple baseline that predicts the class distribution for each shot (i.e., the proportion of shots yielding a goal: 10.51\%)~\cite{decroos2019interpretable}.
\end{description}

We split the dataset into a training set and a test set. The training set spans the 2017/2018 through 2019/2020 seasons and includes 136,769 shots. The test set covers the 2020/2021 season and includes 38,737 shots.

We train Explainable Boosting Machine models for the first three approaches using the \textsc{InterpretML} Python package, which does not require extensive hyperparameter tuning. Based on limited experimentation on the training set, we let the learning algorithm use maximum 64 bins for the feature functions and 32 bins for the pairwise feature functions. We use the default values for the remaining hyperparameters. The learning algorithm uses 15\% of the available training data as a validation set to prevent over-fitting in the boosting step.

We compute seven performance metrics for probabilistic classification. We compute the widely-used AUC-ROC, Brier score (BS) and logarithmic loss (LL) performance metrics~\cite{ferri2009experimental}. Like~\cite{decroos2019interpretable}, we compute normalized variants of both Brier score (NBS) and logarithmic loss (NLL), where the score for each approach is divided by the score for the baseline approach. In addition, we compute the expected calibration error (ECE)\footnote{\url{https://github.com/ML-KULeuven/soccer\_xg}} and a normalized variant thereof (NECE)~\cite{guo2017calibration}. For AUC-ROC, a higher score is better. For all other performance metrics, a lower score is better.

\subsection{Performance Comparison}
\label{subsec:experimental-evaluation-comparison}

\begin{table*}[ht]
    \centering
    \begin{tabular}{l||r||rr||rr||rr}
        \toprule
        \textbf{Approach} & \textbf{AUC-ROC} & \textbf{BS} & \textbf{NBS} & \textbf{LL} & \textbf{NLL} & \textbf{ECE} & \textbf{NECE} \\
        \midrule
        Soft zones & 0.7922 & 0.0825 & 0.8120 & \textbf{0.2869} & \textbf{0.8043} & \textbf{0.0020} & \textbf{0.2105} \\
        Hard zones & 0.7743 & 0.0838 & 0.8248 & 0.2930 & 0.8214 & 0.0027 & 0.2842 \\
        Distance and angle & \textbf{0.7932} & \textbf{0.0823} & \textbf{0.8100} & 0.2878 & 0.8068 & 0.0025 & 0.2632 \\
        Naive baseline & 0.5000 & 0.1016 & 1.0000 & 0.3567 & 1.0000 & 0.0095 & 1.0000 \\
        \bottomrule
    \end{tabular}
    \caption{AUC-ROC, Brier score (BS), normalized Brier score (NBS), logarithmic loss (LL), normalized logarithmic loss (NLL), expected calibration error (ECE) and normalized expected calibration error (NECE) for the four considered approaches. For AUC-ROC, a higher score is better. For all other performance metrics, a lower score is better. The best result for each metric is in bold. The proposed \textit{Soft zones} approach yields comparable performance to the traditional \textit{Distance and angle} approach, while the estimates are easier to explain to practitioners.}
    \label{tbl:results}
\end{table*}

\begin{figure*}[ht]
    \centering
	\begin{subfigure}[b]{0.475\textwidth}
        \centering
        \includegraphics[width=\linewidth]{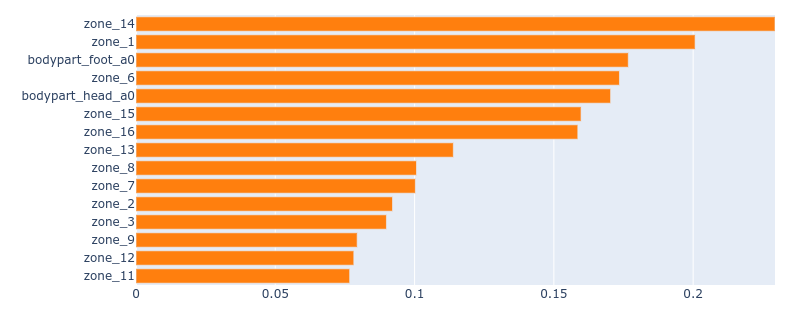}
        \caption{Overall importance of each feature towards making predictions about shots. The vertical axis presents the different features, whereas the horizontal axis reports the overall importance of each feature, where a higher score corresponds to a higher importance.}
        \label{fig:analysis-summary}
    \end{subfigure}
    \hfill
    \begin{subfigure}[b]{0.475\textwidth}
        \centering 
        \includegraphics[width=\linewidth]{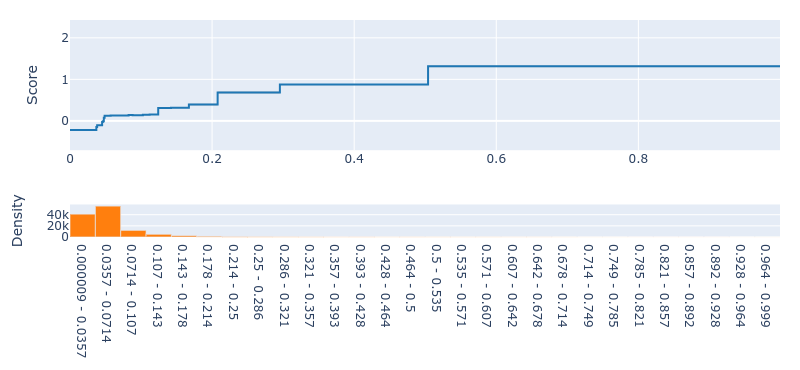}
        \caption{Impact of the feature value for the feature that corresponds to the zone right in front of goal (\textbf{zone\_1}) on the prediction. The score increases as the value for this feature increases, where the score is expressed in log odds.}
        \label{fig:analysis-zone1}
    \end{subfigure}
    \vskip\baselineskip
    \begin{subfigure}[b]{0.475\textwidth}
        \centering 
        \includegraphics[width=\linewidth]{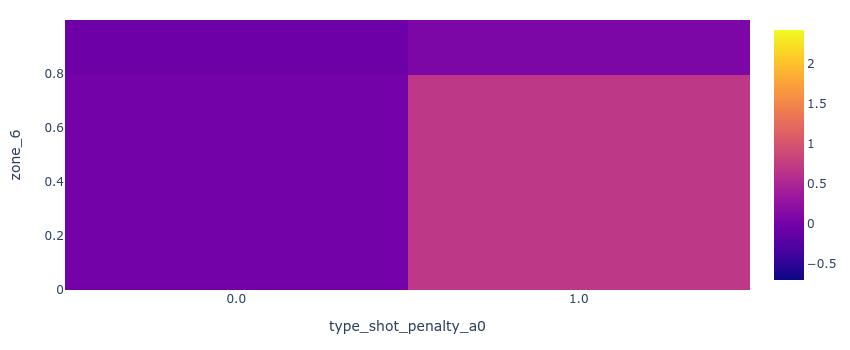}
        \caption{Pairwise interaction between the zone that contains the penalty spot (\textbf{zone\_6}) and the indicator whether the shot is a penalty or not. Surprisingly, the score drops for shots close to the penalty spot (i.e., high value for \textbf{zone\_6}) in case the shot is a penalty (i.e., \textbf{type\_shot\_penalty\_a0}) is 1.}
        \label{fig:analysis-penalty-zone6}
    \end{subfigure}
    \hfill
    \begin{subfigure}[b]{0.475\textwidth}
        \centering 
        \includegraphics[width=\linewidth]{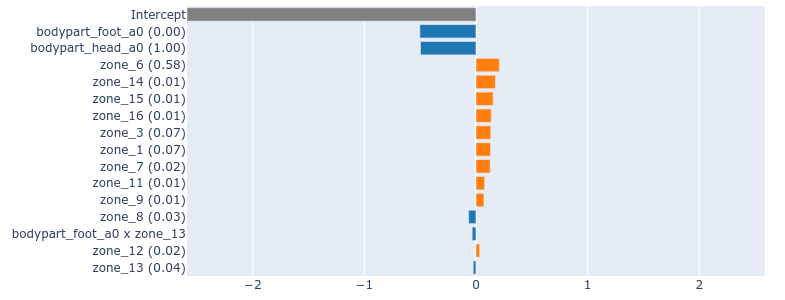}
        \caption{Explanation for a random shot in the test set. The vertical axis presents the different features, whereas the horizontal axis reports the overall importance of each feature, where a higher score corresponds to a higher importance.}
        \label{fig:analysis-shot}
    \end{subfigure}
    \caption{Insights into the learned model (\ref{fig:analysis-summary},~\ref{fig:analysis-zone1} and~\ref{fig:analysis-penalty-zone6}) as well as the prediction for a random shot in the test set (\ref{fig:analysis-shot}).} 
    \label{fig:analysis}
\end{figure*}

Table~\ref{tbl:results} shows the AUC-ROC, Brier score (BS), normalized Brier score (NBS), logarithmic loss (LL), normalized logarithmic loss (NLL), expected calibration error (ECE) and normalized expected calibration error (NECE) for all four approaches on the test set. The best result for each of the performance metrics is highlighted in bold.

The proposed \textit{Soft zones} approach clearly outperforms the \textit{Hard zones} approach across all metrics and yields similar performance to the traditional \textit{Distance and angle} approach. Moreover, the performance of the proposed \textit{Soft zones} approach is comparable to the performance of other approaches on similar datasets~\cite{robberechts2020data}.

Since the probability estimates that the models produce should be explainable to soccer practitioners, the models should not only be accurate but also robust against adversarial examples. It would be challenging to explain to soccer practitioners why a model produces vastly different probability estimates for two shots performed under very similar circumstances. Research by~\cite{devos2020versatile} and~\cite{devos2021verifying} suggests that the Gradient Boosted Decision Tree models used by the reference implementation of the VAEP framework~\cite{decroos2019actions} in the \textsc{socceraction} Python package\footnote{\url{https://github.com/ML-KULeuven/socceraction}} are vulnerable to adversarial examples, which is undesirable for our use case. Early experiments suggest that the Explainable Boosting Machine models are less vulnerable to adversarial examples but further research is required.

\subsection{Explanation Analysis}
\label{subsec:experimental-evaluation-analysis}

Figure~\ref{fig:analysis} shows insights into the learned model (\ref{fig:analysis-summary},~\ref{fig:analysis-zone1} and~\ref{fig:analysis-penalty-zone6}) as well as the prediction for a random shot in the test set (\ref{fig:analysis-shot}).

Figure~\ref{fig:analysis-summary} shows the overall importance of each feature towards making predictions about shots. The vertical axis presents the different features, whereas the horizontal axis reports the overall importance of each feature, where a higher score corresponds to a higher importance. The first-ranked feature corresponds to the zone furthest away from the goal (\textbf{zone\_14}). Shots from this zone rarely result in a goal. In contrast, the second-ranked feature corresponds to the zone right in front of goal (\textbf{zone\_1}). Shots from this zone often result in a goal. Hence, both features are useful to make predictions about the outcome of a shot.

Figure~\ref{fig:analysis-zone1} shows the impact of the feature value for the feature that corresponds to the zone right in front of goal (\textbf{zone\_1}) on the prediction. The score increases as the value for this feature increases, where the score is expressed in log odds. Intuitively, the prediction for a shot increases as the shot is closer to the center of this zone.

Figure~\ref{fig:analysis-penalty-zone6} shows the pairwise interaction between the zone that contains the penalty spot (\textbf{zone\_6}) and the indicator whether the shot is a penalty or not. Surprisingly, the score drops for shots close to the penalty spot in case the shot is a penalty. This behavior is likely caused by the presence of a separate feature function that captures whether the shot is a penalty or not.  To improve the explainability of the model, either this pairwise feature function or the feature function corresponding to the penalty indicator could be dropped.

Figure~\ref{fig:analysis-shot} shows the explanation for a random shot in the test set. The vertical axis presents the different features and the corresponding values for the shot, whereas the horizontal axis reports the overall importance of each feature, where a higher score corresponds to a higher importance. The shot is headed (i.e.,~\textbf{bodypart\_head\_a0} is 1) instead of footed (i.e., \textbf{bodypart\_foot\_a0} is 0), which has a negative impact on the prediction. In contrast, the shot is taken from close to the center of the zone that includes the penalty spot (i.e., \textbf{zone\_6} is 0.58), which has a positive impact on the prediction.

\section{Related Work}
\label{sec:related-work}

The task of estimating the probability of a shot resulting in a goal has received a lot of attention in the soccer analytics community, both inside and outside\footnote{\url{https://wikieducator.org/Sport_Informatics_and_Analytics/Performance_Monitoring/Expected_Goals}} the academic literature. The earliest mentions of ``expected-goals models'' date back to around 2013 when soccer analytics enthusiasts started describing their approaches in blog posts~\cite{caley2013shot,ijtsma2013forget,ijtsma2013where,eastwood2014expected2,eastwood2014expected,ijtsma2015close,eastwood2015expected,caley2015premier}. Most approaches first define a number of shot characteristics and then fit a Logistic Regression model. The research primarily focuses on identifying shot characteristics that are predictive of the outcome of a shot. While these blog posts carry many useful insights, the experimental evaluation is often rather limited~\cite{mackay2016how}. Unlike the non-academic literature, the limited academic literature in this area primarily focuses on approaches that use more sophisticated machine learning techniques and operate on tracking data~\cite{lucey2014quality,anzer2021goal}.

Due to the wider availability of finer-grained data, the broader task of valuing on-the-ball player actions in matches has received a lot of attention in recent years. The low-scoring nature of soccer makes the impact of a single action hard to quantify. Therefore, researchers have proposed many different approaches to distribute the reward for scoring a goal across the actions that led to the goal-scoring opportunity, including the shot itself. Some of these approaches operate on match event data~\cite{rudd2011framework,decroos2017starss,decroos2019actions,decroos2019interpretable,vanroy2020valuing,liu2020deep}, whereas other approaches operate on tracking data~\cite{link2016real,spearman2018beyond,dick2019learning,fernandez2019decomposing}.

The work by \cite{decroos2019interpretable} comes closest to our work. They replace the Gradient Boosted Decision Tree models in the reference implementation of the VAEP framework~\cite{decroos2019actions} with Explainable Boosting Machine models that are easier to interpret. Unlike our approach, their approach uses the original VAEP features, some of which are challenging to explain to soccer practitioners who are unfamiliar with mathematics and machine learning.

\section{Conclusion and Future Work}
\label{sec:conclusion-future-work} 

This paper addresses the task of estimating the probability that a shot will result in a goal such that the estimates can be explained to soccer practitioners who are unfamiliar with mathematics and machine learning. Inspired by ~\cite{decroos2019interpretable}, we train an Explainable Boosting Machine model that can easily be inspected and interpreted due to its additive nature. We represent each shot location by a feature vector that reflects a fuzzy assignment of a shot to designated zones on the pitch that soccer practitioners use to convey tactical concepts and insights about the game. In contrast, existing approaches use more complicated location representations that are designed to facilitate the learning process but are less straightforward to explain to soccer practitioners. In addition to being easier to explain to practitioners, our approach exhibits similar performance to existing approaches.

In the future, we aim to improve and extend our approach. One research direction is to extend the approach from valuing shots to valuing other types of actions such as passes, crosses and dribbles. Another research direction is to refine the zones for representing shot locations in collaboration with soccer practitioners. Yet another research direction is to investigate whether the proposed location representation performs well for Gradient Boosted Decision Tree models, which are widely used in soccer analytics nowadays. Furthermore, we aim to perform more thorough comparative tests with soccer practitioners and to investigate whether the proposed location representation leads to more robust models that are less vulnerable to adversarial examples. 



\balance
\bibliographystyle{named}
\bibliography{ijcai21}

\begin{thebibliography}{}

\bibitem[\protect\citeauthoryear{Anzer and Bauer}{2021}]{anzer2021goal}
Gabriel Anzer and Pascal Bauer.
\newblock {A Goal Scoring Probability Model for Shots Based on Synchronized
  Positional and Event Data in Football (Soccer)}.
\newblock {\em Frontiers in Sports and Active Living}, 3, 2021.

\bibitem[\protect\citeauthoryear{Bezdek}{1981}]{bezdek1981pattern}
James Bezdek.
\newblock {\em {Pattern Recognition with Fuzzy Objective Function Algorithms:
  Advanced Applications in Pattern Recognition}}.
\newblock Springer, 1981.

\bibitem[\protect\citeauthoryear{Bransen and
  Van~Haaren}{2020}]{bransen2020player}
Lotte Bransen and Jan Van~Haaren.
\newblock {Player Chemistry: Striving for a Perfectly Balanced Soccer Team}.
\newblock In {\em Proceedings of the 14th MIT Sloan Sports Analytics
  Conference}, 2020.

\bibitem[\protect\citeauthoryear{Bransen \bgroup \em et al.\egroup
  }{2019}]{bransen2019choke}
Lotte Bransen, Pieter Robberechts, Jan Van~Haaren, and Jesse Davis.
\newblock {Choke or Shine? Quantifying Soccer Players' Abilities to Perform
  Under Mental Pressure}.
\newblock In {\em Proceedings of the 13th MIT Sloan Sports Analytics
  Conference}, 2019.

\bibitem[\protect\citeauthoryear{Caley}{2013}]{caley2013shot}
Michael Caley.
\newblock {Shot Matrix I: Shot Location and Expected Goals}, 2013.

\bibitem[\protect\citeauthoryear{Caley}{2015}]{caley2015premier}
Michael Caley.
\newblock {Premier League Projections and New Expected Goals}, 2015.

\bibitem[\protect\citeauthoryear{Decroos and
  Davis}{2019}]{decroos2019interpretable}
Tom Decroos and Jesse Davis.
\newblock {Interpretable Prediction of Goals in Soccer}.
\newblock In {\em Proceedings of the AAAI-20 Workshop on Artificial
  Intelligence in Team Sports}, 2019.

\bibitem[\protect\citeauthoryear{Decroos \bgroup \em et al.\egroup
  }{2017}]{decroos2017starss}
Tom Decroos, Jan Van~Haaren, Vladimir Dzyuba, and Jesse Davis.
\newblock {STARSS: A Spatio-Temporal Action Rating System for Soccer}.
\newblock In {\em International Workshop on Machine Learning and Data Mining
  for Sports Analytics}, pages 11--20, 2017.

\bibitem[\protect\citeauthoryear{Decroos \bgroup \em et al.\egroup
  }{2019}]{decroos2019actions}
Tom Decroos, Lotte Bransen, Jan Van~Haaren, and Jesse Davis.
\newblock {Actions Speak Louder than Goals: Valuing Player Actions in Soccer}.
\newblock In {\em Proceedings of the 25th ACM SIGKDD International Conference
  on Knowledge Discovery and Data Mining}, 2019.

\bibitem[\protect\citeauthoryear{Decroos \bgroup \em et al.\egroup
  }{2020}]{decroos2020soccermix}
Tom Decroos, Maaike Van~Roy, and Jesse Davis.
\newblock {SoccerMix: Representing Soccer Actions with Mixture Models}.
\newblock In {\em Proceedings of the 2020 European Conference on Machine
  Learning and Principles and Practice of Knowledge Discovery in Databases},
  2020.

\bibitem[\protect\citeauthoryear{Devos \bgroup \em et al.\egroup
  }{2020}]{devos2020versatile}
Laurens Devos, Wannes Meert, and Jesse Davis.
\newblock {Versatile Verification of Tree Ensembles}, 2020.

\bibitem[\protect\citeauthoryear{Devos \bgroup \em et al.\egroup
  }{2021}]{devos2021verifying}
Laurens Devos, Wannes Meert, and Jesse Davis.
\newblock {Verifying Tree Ensembles by Reasoning about Potential Instances}.
\newblock In {\em Proceedings of the 2021 SIAM International Conference on Data
  Mining}, pages 450--458, 2021.

\bibitem[\protect\citeauthoryear{Dick and Brefeld}{2019}]{dick2019learning}
Uwe Dick and Ulf Brefeld.
\newblock {Learning to Rate Player Positioning in Soccer}.
\newblock {\em Big Data}, 7(1):71--82, 2019.

\bibitem[\protect\citeauthoryear{Dunn}{1973}]{dunn1973fuzzy}
Joseph Dunn.
\newblock {A Fuzzy Relative of the ISODATA Process and Its Use in Detecting
  Compact Well-Separated Clusters}.
\newblock {\em Journal of Cybernetics}, 3(3), 1973.

\bibitem[\protect\citeauthoryear{Eastwood}{2014a}]{eastwood2014expected2}
Martin Eastwood.
\newblock {Expected Goals: Foot Shots Versus Headers}, 2014.

\bibitem[\protect\citeauthoryear{Eastwood}{2014b}]{eastwood2014expected}
Martin Eastwood.
\newblock {Expected Goals For All}, 2014.

\bibitem[\protect\citeauthoryear{Eastwood}{2015}]{eastwood2015expected}
Martin Eastwood.
\newblock {Expected Goals and Support Vector Machines}, 2015.

\bibitem[\protect\citeauthoryear{Fernández \bgroup \em et al.\egroup
  }{2019}]{fernandez2019decomposing}
Javier Fernández, Luke Bornn, and Dan Cervone.
\newblock {Decomposing the Immeasurable Sport: A Deep Learning Expected
  Possession Value Framework for Soccer}.
\newblock In {\em Proceedings of the 13th MIT Sloan Sports Analytics
  Conference}, 2019.

\bibitem[\protect\citeauthoryear{Ferri \bgroup \em et al.\egroup
  }{2009}]{ferri2009experimental}
César Ferri, José Hernández-Orallo, and R.~Modroiu.
\newblock {An Experimental Comparison of Performance Measures for
  Classification}.
\newblock {\em Pattern Recognition Letters}, 30(1), 2009.

\bibitem[\protect\citeauthoryear{Guo \bgroup \em et al.\egroup
  }{2017}]{guo2017calibration}
Chuan Guo, Geoff Pleiss, Yu~Sun, and Kilian Weinberger.
\newblock {On Calibration of Modern Neural Networks}.
\newblock In {\em International Conference on Machine Learning}, 2017.

\bibitem[\protect\citeauthoryear{Ijtsma}{2013a}]{ijtsma2013forget}
Sander Ijtsma.
\newblock {Forget Shot Numbers, Let’s Use Expected Goals Instead}, 2013.

\bibitem[\protect\citeauthoryear{Ijtsma}{2013b}]{ijtsma2013where}
Sander Ijtsma.
\newblock {Where Do the Best Shots Come From?}, 2013.

\bibitem[\protect\citeauthoryear{Ijtsma}{2015}]{ijtsma2015close}
Sander Ijtsma.
\newblock {A Close Look at My New Expected Goals Model}, 2015.

\bibitem[\protect\citeauthoryear{Link \bgroup \em et al.\egroup
  }{2016}]{link2016real}
Daniel Link, Steffen Lang, and Philipp Seidenschwarz.
\newblock {Real Time Quantification of Dangerousity in Football Using
  Spatiotemporal Tracking Data}.
\newblock {\em PloS One}, 11(12):e0168768, 2016.

\bibitem[\protect\citeauthoryear{Liu \bgroup \em et al.\egroup
  }{2020}]{liu2020deep}
Guiliang Liu, Yudong Luo, Oliver Schulte, and Tarak Kharrat.
\newblock {Deep Soccer Analytics: Learning an Action-Value Function for
  Evaluating Soccer Players}.
\newblock {\em Data Mining and Knowledge Discovery}, 34(5):1531--1559, 2020.

\bibitem[\protect\citeauthoryear{Lucey \bgroup \em et al.\egroup
  }{2014}]{lucey2014quality}
Patrick Lucey, Alina Bialkowski, Mathew Monfort, Peter Carr, and Iain Matthews.
\newblock {Quality vs Quantity: Improved Shot Prediction in Soccer Using
  Strategic Features from Spatiotemporal Data}.
\newblock In {\em Proceedings of the 8th MIT Sloan Sports Analytics
  Conference}, 2014.

\bibitem[\protect\citeauthoryear{Mackay}{2016}]{mackay2016how}
Nils Mackay.
\newblock {How Not to Evaluate Your xG Model}, 2016.

\bibitem[\protect\citeauthoryear{Maric}{2014}]{maric2014halfspaces}
René Maric.
\newblock {The Half-Spaces}, 2014.

\bibitem[\protect\citeauthoryear{Nori \bgroup \em et al.\egroup
  }{2019}]{nori2019interpretml}
Harsha Nori, Samuel Jenkins, Paul Koch, and Rich Caruana.
\newblock {InterpretML: A Unified Framework for Machine Learning
  Interpretability}, 2019.

\bibitem[\protect\citeauthoryear{Robberechts and
  Davis}{2020a}]{robberechts2020data}
Pieter Robberechts and Jesse Davis.
\newblock {How Data Availability Affects the Ability to Learn Good xG Models}.
\newblock In {\em International Workshop on Machine Learning and Data Mining
  for Sports Analytics}, pages 17--27, 2020.

\bibitem[\protect\citeauthoryear{Robberechts and
  Davis}{2020b}]{robberechts2020interplay}
Pieter Robberechts and Jesse Davis.
\newblock {Illustrating the Interplay Between Features and Models in xG}, 2020.

\bibitem[\protect\citeauthoryear{Rudd}{2011}]{rudd2011framework}
Sarah Rudd.
\newblock {A Framework for Tactical Analysis and Individual Offensive
  Production Assessment in Soccer Using Markov Chains}.
\newblock In {\em New England Symposium on Statistics in Sports}, 2011.

\bibitem[\protect\citeauthoryear{Spearman}{2018}]{spearman2018beyond}
William Spearman.
\newblock {Beyond Expected Goals}.
\newblock In {\em Proceedings of the 12th MIT Sloan Sports Analytics
  Conference}, 2018.

\bibitem[\protect\citeauthoryear{Van~Roy \bgroup \em et al.\egroup
  }{2020}]{vanroy2020valuing}
Maaike Van~Roy, Pieter Robberechts, Tom Decroos, and Jesse Davis.
\newblock {Valuing On-the-Ball Actions in Soccer: A Critical Comparison of XT
  and VAEP}.
\newblock In {\em Proceedings of the AAAI-20 Workshop on Artifical Intelligence
  in Team Sports}, 2020.

\end{thebibliography}

\end{document}